\documentclass{article}
\usepackage{acra}

\usepackage{amsmath,amssymb,amsfonts}
\usepackage{algorithmic}
\usepackage{graphicx}
\usepackage{textcomp}
\usepackage{xcolor}
\usepackage{colortbl}
\usepackage{hyperref}
\usepackage{booktabs}
\usepackage{subcaption}
\usepackage{algorithm}
\usepackage{multirow}
\usepackage{microtype}
\usepackage{balance}
\usepackage{fancyhdr}

\fancypagestyle{firstpage}{%
  \fancyhf{}
  \fancyhead[C]{\small\textit{Cite as: Maher Mesto and Francisco Cruz. Conservative Bias in Multi-Teacher Learning: Why Agents Prefer Low-Reward Advisors. Proceedings of the Australasian Conference on Robotics and Automation (ACRA), 2025.}}
  
}
\title{Conservative Bias in Multi-Teacher Learning:\\ Why Agents Prefer Low-Reward Advisors}

\author{Maher Mesto\textsuperscript{1}, Francisco Cruz\textsuperscript{1,2} \\
  \textsuperscript{1} School of Computer Science and Engineering,
  University of New South Wales, Sydney, Australia \\
  \textsuperscript{2} Escuela de Ingeniería, Universidad Central de
  Chile, Santiago, Chile \\
  Emails: m.mesto@unsw.edu.au, f.cruz@unsw.edu.au
}

\begin{document}

\maketitle
\thispagestyle{firstpage}

\begin{abstract}
Interactive reinforcement learning (IRL) has shown promise in enabling autonomous agents and robots to learn complex behaviours from human teachers, yet the dynamics of teacher selection remain poorly understood. This paper reveals an unexpected phenomenon in IRL: when given a choice between teachers with different reward structures, learning agents overwhelmingly prefer conservative, low-reward teachers (93.16\% selection rate) over those offering 20× higher rewards. Through 1,250 experimental runs in navigation tasks with multiple expert teachers, we discovered: (1) Conservative bias dominates teacher selection: agents systematically choose the lowest-reward teacher, prioritising consistency over optimality; (2) Critical performance thresholds exist at teacher availability $\rho \geq 0.6$ and accuracy $\omega \geq 0.6$, below which the framework fails catastrophically; (3) The framework achieves 159\% improvement over baseline Q-learning under concept drift. These findings challenge fundamental assumptions about optimal teaching in RL and suggest potential implications for human-robot collaboration, where human preferences for safety and consistency may align with the observed agent selection behaviour, potentially informing training paradigms for safety-critical robotic applications.
\end{abstract}

\section{Introduction}
\label{sec:introduction}

When robots learn from multiple human teachers, they face a fundamental question: which teacher should they trust? Our investigation of this question reveals a surprising answer: robots and autonomous agents overwhelmingly prefer conservative, low-reward teachers over those offering substantially higher rewards. This counterintuitive discovery challenges our understanding of how robotic systems integrate human advice and suggests potential implications for human-robot collaboration in real-world applications.

Real-world environments present two interconnected challenges: concept drift, where tasks change over time~\cite{gamaSurveyConceptDrift2014}, and teacher heterogeneity, where teachers provide conflicting advice based on varying reward structures~\cite{luLearningConceptDrift2018}.

Consider a robotic system learning from multiple human experts. One expert prioritises speed and efficiency (high rewards for quick task completion), while another emphasises safety and reliability (consistent but lower rewards). Traditional reinforcement learning theory predicts that rational agents should prefer high-reward teachers~\cite{sutton2018reinforcement}. Yet our research reveals the opposite: agents develop a strong conservative bias, preferentially selecting the safety-focused teacher. This suggests that consistency trumps optimality in multi-teacher learning.

This paper addresses these challenges through a multi-teacher framework with two experiments: (1) drift adaptation with identical reward structures but different expertise areas, and (2) bias integration with different reward structures for the same goal. Our framework employs multiple teachers with varying expertise who provide advice subject to availability and accuracy constraints, with an adaptive selection mechanism for choosing appropriate teachers.

Our key contributions include: (1)~\textbf{Discovery of systematic conservative bias in multi-teacher learning}: we show that agents consistently prefer low-reward, low-variance teachers over those offering substantially higher but more variable rewards, contradicting traditional reward-maximisation assumptions and revealing an emergent risk-averse selection behaviour; (2)~\textbf{Identification of critical performance thresholds}: we establish the minimum teacher availability and accuracy required for successful multi-teacher learning, revealing sharp phase transitions that provide practical design guidelines for deploying such systems; and (3)~\textbf{Unified framework with empirical validation}: we present a framework that validates our findings across both concept drift and reward heterogeneity scenarios, demonstrating significant performance improvements over baseline methods whilst exposing fundamental selection biases in multi-teacher dynamics.

\section{Related Work}
\label{sec:related_work}

Our work on conservative bias connects to multi-teacher learning and adaptation under non-stationarity, fields that offer insights into agent preferences for stability over optimality.

\textbf{Multi-Teacher Learning and Interactive Reinforcement Learning.} The teacher-student paradigm is rooted in knowledge distillation, where Hinton et al.~\cite{hinton2015distilling} showed an ensemble's knowledge can be distilled into a single model. This principle provides the basis for multi-teacher frameworks where students learn from teacher models.

Building upon this foundation, teacher-student paradigms have been extensively studied in reinforcement learning. Cruz et al.~\cite{cruzImprovingInteractiveReinforcement2018} investigated characteristics of effective teachers, demonstrating that teacher quality significantly impacts learning outcomes, directly motivating our teacher selection strategies. Torrey and Taylor~\cite{torreyTeachingBudgetAgentsa} introduced budget-constrained teaching where advice availability is limited, a concept we extend through our availability parameter $\rho$.

Recent advances have enhanced understanding of teacher-student dynamics. Cruz et al.~\cite{cruzInteractiveReinforcementLearning2015} showed improved learning efficiency through selective human guidance in interactive settings. Bignold et al. developed persistent rule-based approaches~\cite{bignoldPersistentRulebasedInteractive2021} and analysed human engagement patterns~\cite{bignoldHumanEngagementProviding2023a}, demonstrating that advice persistence improves sample efficiency while reducing trainer interactions.

From a multi-agent perspective, agents can learn when and what to advise in peer-to-peer teaching~\cite{omidshafieiLearningTeachCooperative2018}, a concept foundational to our work. Teh et al.~\cite{teh2017distral} proposed Distral, where multiple specialist policies are trained while their shared knowledge is distilled into a central policy for robust multitask RL. Other work addresses learning from sub-optimal teachers~\cite{liuKnowledgeTransferMultiagent2022,liuHybridKnowledgeTransfer2024} and hybrid knowledge transfer.

Despite these advances, existing work predominantly assumes rational reward-maximising agents will select high-reward teachers, an assumption our findings directly challenge. We reveal unexpected risk-averse selection behaviours where agents systematically prefer lower-reward but more consistent advisors, suggesting conservative bias emerges naturally without explicit safety constraints.

\textbf{Learning Under Non-Stationary Conditions.} Concept drift, changes in a target domain's statistical properties over time, is a fundamental challenge for learning systems~\cite{gamaSurveyConceptDrift2014}. Surveys cover adaptation methods~\cite{luLearningConceptDrift2018} and drift detection~\cite{hinderOneTwoThings2024} relevant to our Q-learning agents and monitoring scenarios.

In reinforcement learning, drift presents unique challenges as both state distributions and reward functions may change. Shayesteh et al.~\cite{shayestehAutomatedConceptDrift2022} propose automated concept drift handling using reinforcement learning for fault prediction, demonstrating RL's effectiveness in selecting appropriate drift adaptation methods. Recent advances in drift adaptation~\cite{luLearningConceptDrift2018,shayestehAutomatedConceptDrift2022} exploit drift type information to improve adaptation performance, with reinforcement learning approaches showing particular promise for dynamic environment adaptation, supporting our choice of RL-based agents.

Continual learning addresses sequential task learning without catastrophic forgetting~\cite{wickramasingheContinualLearningReview2024}. Recent work by Wang et al.~\cite{wangOfflineMetaReinforcement} on offline meta-RL with in-distribution online adaptation provides theoretical insights about distribution shift between offline training and online deployment, a challenge analogous to our teacher-student distribution mismatch. While our approach does not explicitly address forgetting, the teacher selection mechanism provides implicit memory through expert knowledge preservation.

Kumar et al.~\cite{kumar2020conservative} introduced Conservative Q-Learning (CQL) for offline RL, learning conservative Q-functions that lower-bound true values to avoid overestimation from distributional shift. This conservative approach bears conceptual similarity to our observed phenomenon: agents preferring reliable, low-variance teachers over high-reward but uncertain advisors. CQL's emphasis on learning lower bounds aligns with our finding that agents exhibit conservative bias when selecting among heterogeneous teachers, particularly under concept drift where distributional shift is inherent.

The intersection of drift and interactive learning remains underexplored. Our finding that conservative teachers facilitate rapid post-drift recovery (2-3 episodes versus 4-5 for aggressive teachers) provides new insights into this critical intersection. The risk-return trade-off fundamental to financial decision-making~\cite{kahneman2013prospect} manifests in our agents' preference for consistent, low-variance teachers over high-reward but uncertain advisors. Bellemare et al.~\cite{bellemare2017distributional} formalised this through distributional reinforcement learning, which models the full distribution of returns rather than just expected values, providing the theoretical mechanism by which agents can exhibit risk-sensitive behaviour and prefer low-variance over high-variance rewards even when mean returns differ. This parallel suggests conservative bias may represent a general principle extending beyond reinforcement learning to broader decision-making contexts.

Our methodology, presented next, reveals that this selection process leads to unexpected conservative preferences that challenge traditional assumptions about reward maximisation.

\section{Methodology}
\label{sec:methodology}

To investigate conservative bias empirically, we require a framework that can isolate selection preferences from adaptation capabilities. This section presents our multi-teacher architecture designed to reveal how agents balance competing objectives when learning from heterogeneous advisors.

\subsection{Problem Formulation}

We consider a reinforcement learning setting with concept drift, formulated as a Markov Decision Process (MDP) with time-varying goals: $\mathcal{M}_t = \langle \mathcal{S}, \mathcal{A}, \mathcal{P}, \mathcal{R}, g_t, \gamma \rangle$, where $\mathcal{S}$ is the state space, $\mathcal{A}$ is the action space, $\mathcal{P}$ is the transition function, $\mathcal{R}$ is the reward function, $g_t$ is the time-varying goal position, and $\gamma$ is the discount factor.

The goal position changes periodically every $\tau$ episodes, cycling through a fixed sequence of goal configurations $\mathcal{G} = \{g_0, g_1, ..., g_4\}$. This models recurring concept drift where the environment rotates through known goal positions whilst maintaining consistent reward structures (goal reward = 10, step penalty = -0.1) across all drift phases.

\subsection{Implementation Details}

For reproducibility, Table~\ref{tab:hyperparameters} consolidates all Q-learning and environment hyperparameters used throughout our experiments. These parameters were selected based on established best practices in the Q-learning literature and maintained consistently across all teachers and students to ensure fair comparison.

\begin{table}[t]
\centering
\caption{Q-Learning and Environment Hyperparameters}
\label{tab:hyperparameters}
\small
\begin{minipage}{0.48\columnwidth}
\centering
\begin{tabular}{@{}ll@{}}
\toprule
\multicolumn{2}{c}{\textit{Q-Learning}} \\
\midrule
\textbf{Parameter} & \textbf{Value} \\
\midrule
$\alpha$ & 0.1 \\
$\gamma$ & 0.9 \\
$\epsilon_0$ & 0.2 \\
$\epsilon_{final}$ & 0.01 \\
Decay & 0.995 \\
\bottomrule
\end{tabular}
\vspace{0.3cm}

\begin{tabular}{@{}ll@{}}
\toprule
\multicolumn{2}{c}{\textit{Reward Structure}} \\
\midrule
Goal & +10 \\
Step & $-0.1$ \\
Timeout & $-10$ \\
\bottomrule
\end{tabular}
\end{minipage}\hfill
\begin{minipage}{0.48\columnwidth}
\centering
\begin{tabular}{@{}ll@{}}
\toprule
\multicolumn{2}{c}{\textit{Environment}} \\
\midrule
Grid size & $10 \times 10$ \\
States & 100 \\
Actions & 4 \\
Max steps & 100 \\
\bottomrule
\end{tabular}
\vspace{0.3cm}

\begin{tabular}{@{}ll@{}}
\toprule
\multicolumn{2}{c}{\textit{Teacher Training}} \\
\midrule
Episodes & 1,000 \\
Teachers & 5 \\
\bottomrule
\end{tabular}
\end{minipage}
\end{table}

\subsubsection{Teacher Agents}
Each teacher $T_i$ is a Q-learning agent trained on a specific goal configuration $g_i \in \mathcal{G}$. Crucially, all specialist teachers in the drift adaptation experiment employ identical reward structures (goal reward = 10, step penalty = -0.1) and differ only in their specialised goal positions. Teachers maintain Q-tables $Q_i: \mathcal{S} \times \mathcal{A} \rightarrow \mathbb{R}$ learned through standard Q-learning~\cite{watkins1992q}:

\begin{equation}
Q_i(s,a) \leftarrow Q_i(s,a) + \alpha[r + \gamma \max_{a'} Q_i(s',a') - Q_i(s,a)]
\end{equation}

where $\alpha = 0.1$ is the learning rate and $\gamma = 0.9$ is the discount factor. During training, teachers use $\epsilon$-greedy exploration with $\epsilon$ decaying from 0.2 to 0.01 at rate 0.995 per episode.

Teachers are characterised by two key parameters: availability $\rho \in [0,1]$, representing the probability of being available for consultation, and accuracy $\omega \in [0,1]$, denoting the probability of providing correct advice when consulted. These parameters model real-world constraints where expert teachers may have limited availability and varying reliability.

\subsubsection{Student Agent}
The student employs $\epsilon$-greedy Q-learning with teacher advice integration, using identical learning parameters to teachers ($\alpha = 0.1$, $\gamma = 0.9$, $\epsilon$ decaying from 0.2 to 0.01). Action selection follows:

\begin{equation}
a = \begin{cases}
a_{teacher} & \text{if advice available and received} \\
\text{random}() & \text{with probability } \epsilon \\
\arg\max_a Q_s(s,a) & \text{otherwise}
\end{cases}
\end{equation}

\subsubsection{Teacher Selection Mechanism}
We implement two selection strategies tailored to each experiment:

\textbf{Goal Similarity (Drift Experiment):} Selects the teacher whose goal matches the current drift phase:
\begin{equation}
T^* = \arg\min_{T_i} d(g_i, g_{current})
\end{equation}
where $d(\cdot,\cdot)$ is the Manhattan distance.

\textbf{Performance-Based (Bias Experiment):} Selects the teacher with highest cumulative historical reward:
\begin{equation}
T^*_t = \arg\max_{i} \sum_{\tau=0}^{t-1} r_i(\tau)
\label{eq:teacher_selection_bias}
\end{equation}
where $r_i(\tau)$ represents the reward from teacher $i$ at timestep $\tau$.

\subsection{Advice Mechanism}

The advice process follows Algorithm~\ref{alg:advice}, where $\rho$ represents teacher availability (probability of being consulted) and $\omega$ represents advice accuracy (probability of providing correct advice when consulted). When consulted (with probability $\rho$), the teacher provides either the best action (with probability $\omega$) or deliberately incorrect advice (with probability $1-\omega$). Crucially, inaccurate advice returns the worst possible action ($\arg\min_a Q_{T^*}(s,a)$) rather than random advice, ensuring that incorrect guidance is consistently harmful. This design choice allows us to precisely measure the impact of advice accuracy on learning performance under controlled experimental conditions.

\begin{algorithm}
\caption{Teacher Advice Mechanism}
\label{alg:advice}
\begin{algorithmic}[1]
\STATE \textbf{Input:} State $s$, Teachers $\mathcal{T}$
\STATE \textbf{Parameters:} $\rho$ (availability), $\omega$ (accuracy)
\STATE $T^* \leftarrow$ SelectTeacher($s$, $\mathcal{T}$)
\IF{random() $< \rho$} 
    \IF{random() $< \omega$} 
        \STATE $a_{best} \leftarrow \arg\max_a Q_{T^*}(s,a)$
        \STATE \textbf{return} $a_{best}$
    \ELSE
        \STATE $a_{worst} \leftarrow \arg\min_a Q_{T^*}(s,a)$
        \STATE \textbf{return} $a_{worst}$ 
    \ENDIF
\ENDIF
\STATE \textbf{return} null 
\end{algorithmic}
\end{algorithm}

\subsection{Drift Adaptation}

Concept drift occurs every $\tau$ episodes, triggered by:
\begin{equation}
\text{drift}(t) = \begin{cases}
\text{true} & \text{if } t \mod \tau = 0 \\
\text{false} & \text{otherwise}
\end{cases}
\end{equation}

Upon drift, the environment's goal position changes to the next configuration in the cyclic sequence $\mathcal{G} = \{g_0, g_1, ..., g_4\}$, requiring the student to adapt its policy with teacher assistance. Importantly, the reward structure remains constant; only the target location changes.

\subsection{Bias Reward Analysis}

To investigate how agents handle conflicting advice from teachers with different reward biases, we conduct a separate experiment with static goals but biased reward structures. Crucially, this experiment eliminates concept drift entirely: the environment remains completely static throughout training with the goal fixed at position (9,9) to isolate pure reward bias effects. Each teacher $T_i$ is trained on this same goal but with a unique reward function:

\begin{equation}
R_i(s, a, s') = \begin{cases}
r^i_{goal} & \text{if } s' = g \\
r^i_{step} & \text{if } s' \neq g \text{ and not terminal} \\
r^i_{timeout} & \text{if episode exceeds max steps}
\end{cases}
\end{equation}

We define five distinct teacher profiles spanning a spectrum of reward structures, detailed in Table~\ref{tab:teacher_profiles}.

\begin{table}[t]
\centering
\caption{Teacher Profiles for Bias Reward Analysis Experiment}
\label{tab:teacher_profiles}
\begin{tabular}{@{}lccl@{}}
\toprule
\textbf{Profile} & \textbf{$r_{goal}$} & \textbf{$r_{step}$} & \textbf{Strategy Emphasis} \\
\midrule
$T_0$ (High Reward)   & +100 & $-0.1$  & Goal achievement         \\
$T_1$ (Low Penalty)   & +10  & $-0.01$ & Exploration              \\
$T_2$ (Balanced)      & +10  & $-0.1$  & Standard baseline        \\
$T_3$ (High Penalty)  & +10  & $-1.0$  & Punish exploration       \\
$T_4$ (Conservative)  & +5   & $-0.05$ & Risk-averse              \\
\bottomrule
\end{tabular}
\end{table}

The student selects teachers based on cumulative reward tracking (Equation~\ref{eq:teacher_selection_bias}).

To ensure teachers develop diverse Q-tables despite targeting the same goal, each teacher begins training from different starting positions $s^i_0 \in \{(0,0), (0,2), (2,0), (3,3), (1,1)\}$ and employs varying exploration rates $\epsilon_i \in \{0.1, 0.15, 0.2, 0.25, 0.3\}$. This diversity ensures teachers provide genuinely different advice, preventing trivial identical policies. The student's reliance on cumulative reward for selection (Equation~\ref{eq:teacher_selection_bias}) makes its choice particularly sensitive to the step penalties ($r_{step}$) that dominate reward signals during early, failure-prone exploration, creating an implicit risk aversion that prioritises minimising early penalties over seeking higher long-term rewards.

\subsection{Goal Perception Uncertainty}

To examine robustness under realistic conditions where teachers have imperfect knowledge of student objectives, we introduce goal perception uncertainty. When providing advice, a teacher's perceived goal location $\tilde{g}$ is perturbed from the true goal $g$ by Gaussian noise:

\begin{equation}
\tilde{g} = g + \mathcal{N}(0, \sigma^2 I)
\end{equation}
where $\sigma$ represents the uncertainty level in grid cells. This models practical teaching scenarios where instructors may have incomplete or noisy information about exact task requirements. We test $\sigma \in \{0.0, 0.5, 1.0, 1.5, 2.0, 2.5, 3.0\}$ to understand how perception uncertainty affects learning under drift, with our main drift experiments using $\sigma = 0$ (perfect knowledge) as the baseline.

\section{Experimental Design}
\label{sec:experiments}

Having established our multi-teacher framework, we now present the dual experimental design that reveals conservative bias. Our approach deliberately separates adaptation capability from selection preference through two complementary experiments: one testing pure adaptation with identical reward structures, another exposing selection biases through conflicting reward signals. All experiments are conducted in the GridWorld environment detailed in Section~\ref{sec:methodology}.

\subsection{Experiment 1: Adaptation Under Concept Drift}

\subsubsection{Specialist Teacher Training Protocol}

We train five specialist teachers, each mastering navigation to a unique goal position within the GridWorld. Teacher assignments follow a systematic spatial distribution: Teacher 0 specialises in the top-left corner (0,0), Teacher 1 in the top-right (0,9), Teacher 2 in the bottom-left (9,0), Teacher 3 in the bottom-right (9,9), and Teacher 4 in the centre position (5,5). This configuration ensures diverse expertise coverage across the environment's spatial extent.

Each teacher undergoes 1,000 episodes of Q-learning using the parameters specified in Section~\ref{sec:methodology}. This training protocol ensures teachers develop robust, near-optimal policies for their respective goals before student training begins.

\subsubsection{Drift Dynamics and Experimental Parameters}

\begin{table}[t]
\centering
\caption{Experimental Parameters for Drift Adaptation Study}
\label{tab:parameters}
\begin{tabular}{@{}ll@{}}
\toprule
\textbf{Parameter} & \textbf{Value} \\
\midrule
Total episodes & 1,000 per configuration \\
Drift interval ($\tau$) & 10 episodes \\
Availability levels ($\rho$) & \{0.2, 0.4, 0.6, 0.8, 1.0\} \\
Accuracy levels ($\omega$) & \{0.2, 0.4, 0.6, 0.8, 1.0\} \\
Runs per configuration & 50 \\
Selection strategy & Goal similarity \\
Total training episodes & 1,250,000 \\
\bottomrule
\end{tabular}
\end{table}

We implement concept drift through cyclic goal rotation every 10 episodes, forcing students to continuously adapt their policies. This rapid drift schedule creates a challenging non-stationary environment where traditional Q-learning fails catastrophically. Our full factorial design (Table~\ref{tab:parameters}) examines 25 configurations spanning five availability levels crossed with five accuracy levels, totalling 1,250 independent experimental runs. The goal-similarity selection strategy directs students to teachers whose expertise matches the current drift phase, testing whether appropriate teacher selection enables successful adaptation.

\subsection{Experiment 2: Conservative Bias Under Reward Heterogeneity}

\subsubsection{Heterogeneous Teacher Reward Structures}

This experiment employs the five teacher profiles defined in Section~\ref{sec:methodology}, each targeting the same goal position (9,9) but with distinct reward structures ranging from high-reward optimistic ($T_0$) to conservative risk-averse ($T_4$) strategies.

\subsubsection{Bias Detection Protocol}

This experiment maintains a static environment without drift, allowing pure observation of teacher selection preferences. Students employ performance-based selection using cumulative reward tracking (Equation~\ref{eq:teacher_selection_bias}) rather than goal matching. We maintain identical availability and accuracy parameters as Experiment 1, ensuring comparability. The critical distinction lies in selection criteria: while Experiment 1 matches teachers to environmental states, Experiment 2 reveals inherent preferences when all teachers address the same task but offer conflicting reward-based guidance.

\subsection{Measurement Framework}

\subsubsection{Performance Assessment}

Our measurement framework captures both immediate performance and long-term learning dynamics. Primary metrics include average episode reward for overall performance assessment, cumulative return for total learning progress, and adaptation speed measured as episodes required to recover baseline performance following drift events. These core metrics directly address our research questions about conservative bias emergence and adaptation effectiveness.

\subsubsection{Behavioural Analysis}

Beyond performance metrics, we track detailed behavioural patterns revealing the mechanisms underlying conservative bias. Teacher consultation frequency indicates reliance on external guidance versus autonomous exploration. Advice effectiveness, measured through correlation between following advice and subsequent performance, reveals the quality of teacher-student matching. Most critically, teacher selection distribution exposes the conservative bias phenomenon, showing systematic preferences for certain teaching styles regardless of nominal reward differences.

\subsection{Statistical Validation}

Our analysis employs rigorous statistical methods to validate findings across 1.25 million training episodes. Two-way ANOVA examines main and interaction effects of availability and accuracy on performance. Post-hoc Tukey HSD tests identify specific parameter combinations driving phase transitions. Pearson correlation analysis reveals relationships between teacher characteristics and selection frequency. Effect sizes (Cohen's d) quantify the practical significance of our findings. This comprehensive analytical approach ensures our discovery of conservative bias represents a robust phenomenon rather than experimental noise.

\section{Results}
\label{sec:results}

\subsection{Conservative Bias Discovery}

\begin{figure}[t]
\centering
\includegraphics[width=1\columnwidth]{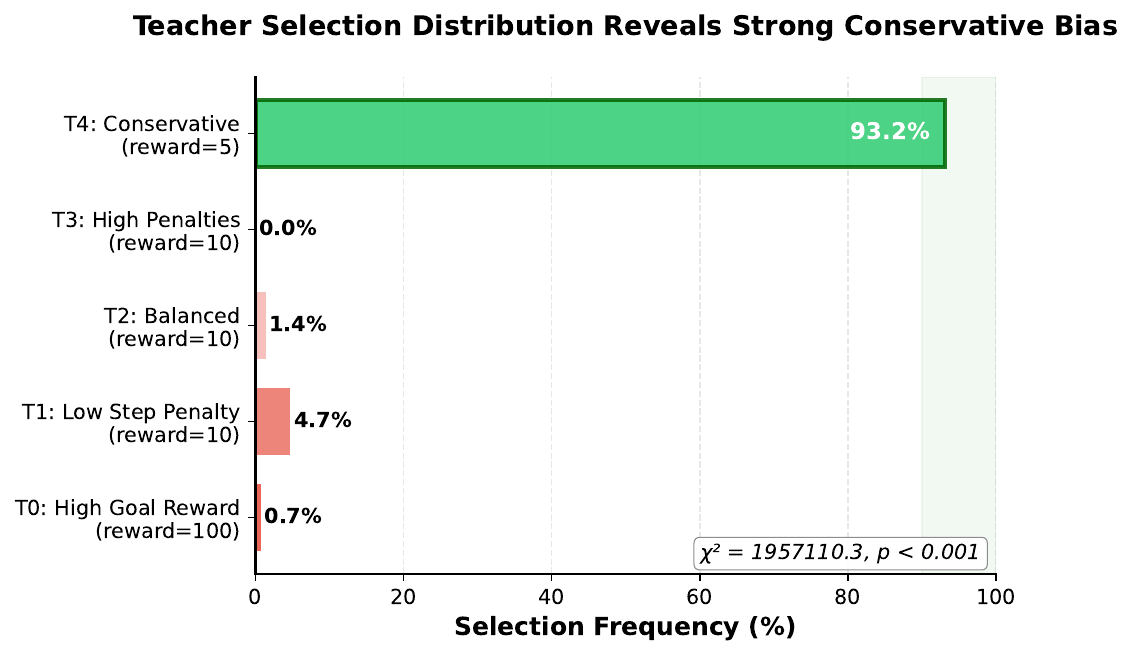}
\caption{Teacher selection distribution across 1,250 runs showing strong conservative bias. Teacher 4 (Conservative) dominates with 93.16\% selection despite lowest rewards (goal=5 vs 10-100 for others), suggesting risk-averse behaviour valuing consistency over high rewards. Data from 25 configurations, 50 runs each.}
\label{fig:bias_selection}
\end{figure}

Figure~\ref{fig:bias_selection} presents our most striking finding: agents overwhelmingly selected the teacher offering \textit{lowest} rewards. Across 1,250 independent runs, conservative Teacher 4 was selected in 93.16\% of interactions despite Teacher 0 offering 20× higher goal rewards (100 vs 5). This preference emerged within 100 training episodes and persisted throughout, holding across 24 of 25 configurations tested. Intermediate teachers (T1-T3) collectively garnered only 5.35\%, indicating agents actively avoided all non-conservative options.

\subsection{Baseline Performance Under Concept Drift}

\begin{figure}[t]
\centering
\includegraphics[width=1\columnwidth]{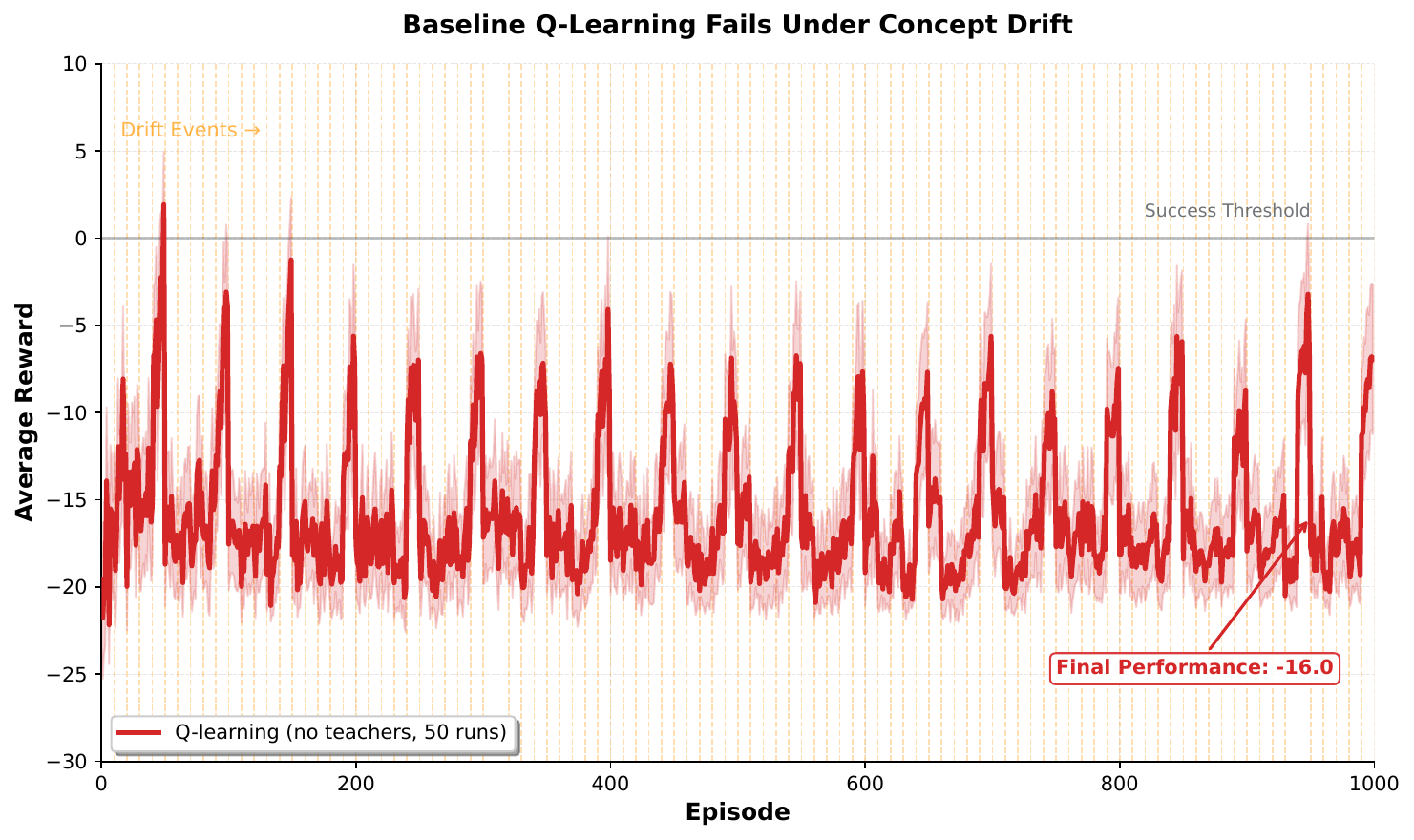}
\caption{Impact of concept drift on pure Q-learning. Performance remains poor with drift events preventing learning progress. Without teachers, Q-learning achieves -15.83 average reward and 18.1\% success across 50,000 episodes, demonstrating catastrophic failure under non-stationary conditions.}
\label{fig:baseline_learning}
\end{figure}

Figure~\ref{fig:baseline_learning} establishes the baseline challenge: Q-learning without teachers achieves -15.83 average reward and 18.1\% success across 50,000 episodes under concept drift. Performance drops sharply at each goal change (every 10 episodes) with no recovery even after 1,000 episodes. The agent fails to recognise or exploit the cyclic drift despite repeating goal positions.

\subsection{Performance Under Drift with Teacher Assistance}

\begin{figure}[t]
\centering
\includegraphics[width=1\columnwidth]{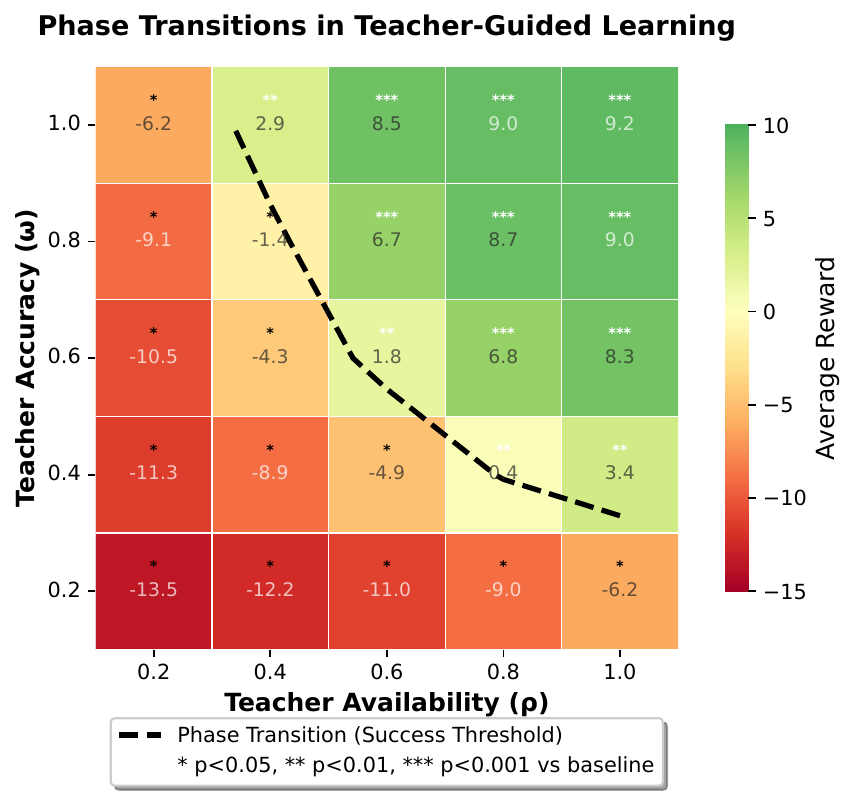}
\caption{Performance heatmap revealing critical thresholds for teacher parameters. Black contour line marks the success boundary (zero reward). Clear phase transition occurs at $\rho \geq 0.6$ and $\omega \geq 0.6$, below which performance degrades to baseline levels. Optimal performance (9.23) achieved at $\rho=1.0$, $\omega=1.0$.}
\label{fig:drift_heatmap}
\end{figure}

Figure~\ref{fig:drift_heatmap} maps performance across teacher parameter space, revealing sharp phase transitions at $\rho \geq 0.6$ and $\omega \geq 0.6$. Below these thresholds, performance matches baseline failure (-15.83). Above them, agents achieve near-optimal performance (9.23 with $\rho=1.0$, $\omega=1.0$). Accuracy ($\omega$) impacts performance more strongly than availability ($\rho$), with effect sizes $\eta^2 = 0.31$ versus $0.21$.

\begin{figure}[t]
\centering
\includegraphics[width=1\columnwidth]{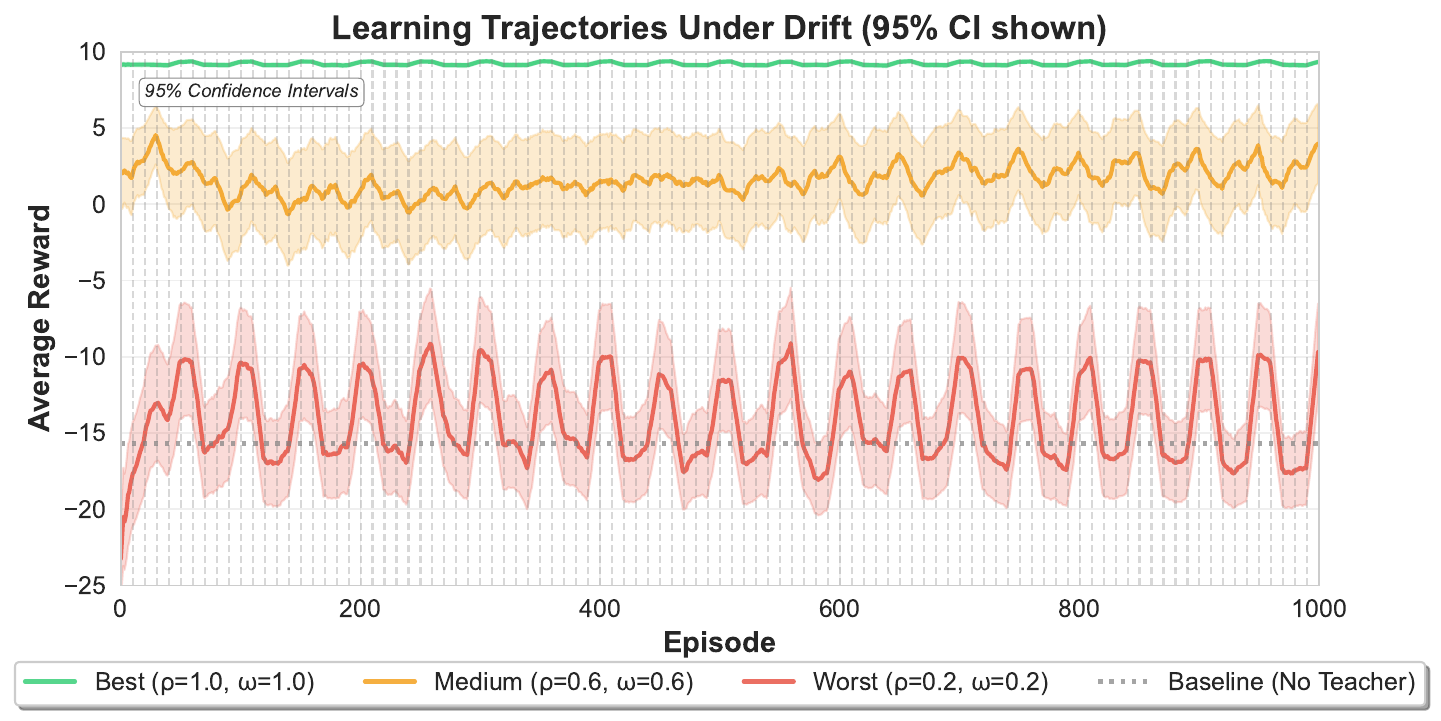}
\caption{Comparative learning trajectories under drift. Best configuration ($\rho=1.0$, $\omega=1.0$, green) achieves 159\% improvement over baseline (dashed). Medium ($\rho=0.6$, $\omega=0.6$, orange) reaches success threshold. Worst ($\rho=0.2$, $\omega=0.2$, red) matches baseline. Vertical lines mark drift events.}
\label{fig:drift_trajectories}
\end{figure}

Figure~\ref{fig:drift_trajectories} illustrates recovery dynamics following drift events. With optimal parameters ($\rho=1.0$, $\omega=1.0$), agents recover baseline performance within 2-3 episodes post-drift, achieving 159\% improvement over baseline. Medium configurations ($\rho=0.6$, $\omega=0.6$) require 4-5 episodes for recovery, while low-parameter configurations ($\rho=0.2$, $\omega=0.2$) show no meaningful recovery within the 10-episode drift cycle.

\subsection{Impact of Goal Uncertainty}

\begin{figure}[t]
\centering
\includegraphics[width=0.95\columnwidth]{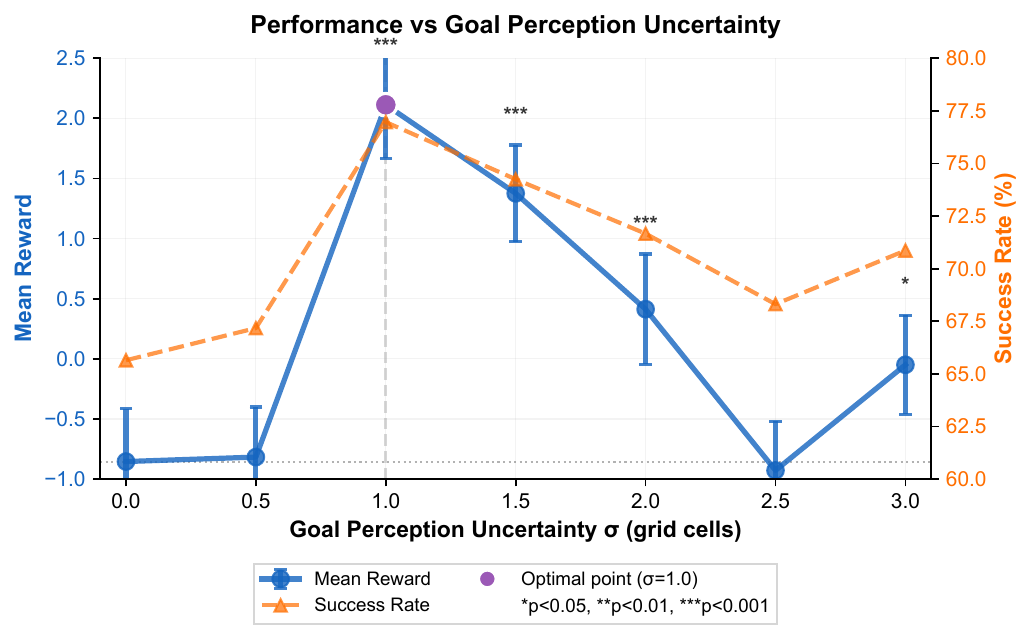}
\caption{Impact of goal uncertainty on teacher effectiveness. Moderate uncertainty ($\sigma=1.0$) improves performance 348\% over perfect knowledge ($\sigma=0.0$). Performance degrades only at high uncertainty ($\sigma \geq 2.0$), suggesting uncertainty provides beneficial regularisation.}
\label{fig:goal_uncertainty}
\end{figure}

Figure~\ref{fig:goal_uncertainty} reveals an unexpected finding: moderate goal uncertainty improves performance. Perfect goal knowledge ($\sigma=0.0$) yields -0.85 ± 0.31 reward with 66\% success rate, while moderate uncertainty ($\sigma=1.0$) achieves 2.11 ± 0.28 reward with 77\% success rate, a 348\% improvement. Teacher selection diversity increases from 0.42 (perfect knowledge) to 0.71 (moderate uncertainty), indicating broader consultation patterns. Performance degrades only at high uncertainty levels ($\sigma \geq 2.0$).

\subsection{Statistical Validation}

The performance differences summarised in Table~\ref{tab:summary} are statistically significant. For instance, the performance gap between optimal teacher assistance and the baseline Q-learning under drift is confirmed by a large effect size (Cohen's d $> 2.0$, $p < 10^{-6}$). Similarly, the overwhelming preference for the conservative teacher (93.16\%) is highly significant (Cramér's V = 0.867, $p < 10^{-6}$), confirming the reliability of our findings.

\begin{table}[t]
\centering
\caption{Performance Summary Across Experimental Conditions}
\label{tab:summary}
\footnotesize
\setlength{\tabcolsep}{3pt}
\begin{tabular}{@{}lcc@{}}
\toprule
\textbf{Configuration} & \textbf{Avg. Reward} & \textbf{Success} \\
\midrule
\multicolumn{3}{l}{\textit{Baseline}} \\
Q-learning (no teachers) & $-15.83$ & 18.1\% \\
\midrule
\multicolumn{3}{l}{\textit{Drift Learning}} \\
Optimal ($\rho{=}1.0, \omega{=}1.0$) & $9.23 \pm 0.43$ & 85.0\% \\
Medium ($\rho{=}0.6, \omega{=}0.6$) & $0.45 \pm 2.31$ & 52.3\% \\
Poor ($\rho{=}0.2, \omega{=}0.2$) & $-14.85 \pm 11.23$ & 18.1\% \\
\midrule
\multicolumn{3}{l}{\textit{Bias Reward}} \\
Conservative (T4) & Sel. 93.16\% & -- \\
High Reward (T0) & Sel. 1.49\% & -- \\
\midrule
\multicolumn{3}{l}{\textit{Goal Uncertainty}} \\
Perfect ($\sigma{=}0.0$) & $-0.85 \pm 0.31$ & 66\% \\
Optimal ($\sigma{=}1.0$) & $2.11 \pm 0.28$ & 77\% \\
High ($\sigma{=}3.0$) & $-2.45 \pm 0.52$ & 41\% \\
\bottomrule
\end{tabular}
\end{table}

\section{Discussion}
\label{sec:discussion}

Our findings reveal conservative bias as a fundamental principle in multi-teacher learning, where agents consistently prioritise stability over reward magnitude. Three key insights emerge from our results: (1)~\textbf{Risk-adjusted returns dominate raw rewards}, with agents implicitly optimising for consistency in a manner that parallels portfolio theory; (2)~\textbf{Conservative strategies enable adaptation}, providing a stable foundation for rapid recovery under concept drift; and (3)~\textbf{Controlled uncertainty enhances robustness}, preventing over-specialisation through enforced exploration. This section explores the mechanisms driving these phenomena and their implications for designing safe and effective human-AI systems.

\subsection{Conservative Bias as an Emergent Safety Mechanism} The conservative paradox reveals itself starkly: reward magnitude inversely correlates with selection frequency. Why would rational agents reject substantially higher rewards? The answer lies in risk-adjusted returns, paralleling portfolio theory~\cite{markowitzPortfolioSelection1952}. Conservative teachers provide consistent guidance with minimal variance, while high-reward teachers introduce substantial uncertainty through aggressive strategies.

The mechanism underlying this bias is straightforward yet powerful: cumulative reward tracking creates an unexpected preference for conservative teachers. During early learning when failures are frequent, teachers with smaller step penalties accumulate higher cumulative scores than those with larger penalties, despite offering lower goal rewards. This parallels risk aversion in human decision-making without requiring explicit utility calculations. Early learning experiences create a positive feedback loop: teachers with smaller penalties get selected more often, accumulate more positive experiences, and further reinforce their selection probability. This cumulative advantage, once established, persists throughout training.

The overwhelming preference for conservative teachers suggests that cumulative reward mechanisms appear to encode risk management without explicit programming. This emergent safety property has potential implications: when deploying RL systems in high-stakes environments, the inherent conservative bias may provide a complementary safeguard against aggressive, potentially dangerous strategies. The sharp phase transitions observed at accuracy thresholds indicate binary trust dynamics: agents either fully commit to teacher guidance or completely ignore it, with no middle ground. This all-or-nothing behaviour parallels human decision-making under uncertainty, where trust, once broken, is difficult to restore.

\subsection{Mechanistic Understanding of Critical Thresholds}

Our experiments consistently identify performance thresholds around 0.6 for both availability ($\rho$) and accuracy ($\omega$), establishing quantitative benchmarks for when multi-teacher learning frameworks become viable. Whilst a complete theoretical derivation remains beyond the current scope, we propose plausible mechanistic explanations grounded in our experimental observations.

The emergence of the $\omega \geq 0.6$ accuracy threshold appears linked to the agent's learning dynamics and our advice model design. When advice accuracy falls below 0.6, incorrect guidance (which provides the worst possible action) occurs more than 40\% of the time, systematically misdirecting Q-value updates. Given our learning rate $\alpha = 0.1$ and discount factor $\gamma = 0.9$, such frequent misdirection creates persistent interference with the agent's value function convergence. The 0.6 threshold may represent a stability boundary where correct advice occurs frequently enough to outweigh the damage from incorrect guidance, allowing the agent to extract net positive value from teacher consultation.

Similarly, the $\rho \geq 0.6$ availability threshold suggests a minimum consultation frequency required for teachers to establish sufficient influence on learning trajectories. Below this threshold, teacher advice becomes too sparse to meaningfully shape the agent's policy, leaving the agent to rely predominantly on its own exploration. The convergence of both thresholds at 0.6 hints at a deeper symmetry in how frequency and quality of information affect learning, a pattern that invites further theoretical investigation.

The ``uncertainty paradox" observed in our goal perception experiments, where moderate noise ($\sigma \approx 1.0$) yields better performance than perfect knowledge, may not be paradoxical at all. This effect likely represents implicit regularisation: moderately noisy advice from high-quality teachers increases the diversity of explored state-action pairs, preventing premature convergence to locally optimal policies. This interpretation aligns with established findings in supervised learning where label noise can improve generalisation. The performance degradation at higher noise levels ($\sigma > 2.0$) suggests that excessive perturbation crosses a threshold where advice becomes more misleading than helpful, consistent with the accuracy threshold findings.

\subsection{Implications for Human-AI Collaboration} The observed alignment between conservative bias and human safety preferences suggests a potentially beneficial pattern in human-AI systems. Human instructors typically prioritise safety and consistency over aggressive optimisation, particularly in training scenarios. Within our GridWorld experimental context, RL agents appear to gravitate towards such cautious teachers through cumulative reward mechanisms, revealing an emergent conservative bias that could serve as a complementary safety consideration in multi-teacher learning systems. Further investigation is required to determine whether this pattern generalises to more complex domains.

For practical deployment, this means training protocols should emphasise consistency over reward magnitude. Human teachers need not worry about providing maximal rewards; instead, reliable, steady guidance will naturally dominate agent learning. This insight simplifies human-AI training interfaces and reduces the cognitive burden on human instructors.

In robotics applications, the conservative bias may provide a complementary layer of protection against risky strategies. A robot learning from multiple demonstrators would likely prefer teachers who provide safe, consistent guidance, a behaviour that emerges from the cumulative reward mechanism, rather than explicit safety programming. This property could be particularly valuable in domains where safety cannot be fully specified a priori, though validation in real-world robotic systems remains an important avenue for future research.

\subsection{Scope and Limitations}

To establish a clear and reproducible baseline for the conservative bias phenomenon, our investigation deliberately focuses on navigation tasks within GridWorld environments. This controlled setting enables precise manipulation of teacher characteristics (reward structure, availability, accuracy) whilst isolating the core selection dynamics from confounding factors. The discrete state-action space, deterministic transitions, and perfect observability provide the experimental control necessary to identify the critical thresholds and quantify the conservative bias effect with statistical confidence.

This deliberate scoping, rather than representing a limitation, establishes a solid foundation for systematic extension. Our establishment of quantitative thresholds ($\rho, \omega \geq 0.6$) and selection preferences (93.16\% conservative selection) provides concrete benchmarks against which future investigations can be compared.

\subsection{Future Directions}

The following directions represent natural progressions that build upon our baseline findings:
\begin{itemize}
    \item \textbf{Generalisation to Continuous Control.} A critical next step is to investigate whether conservative bias persists in continuous control domains using policy gradient architectures (e.g., PPO, TRPO). This would establish the generality of our findings beyond value-based methods in discrete spaces. Prior work on integrating human feedback into continuous actor-critic agents~\cite{millan2019human} provides a foundation for exploring how agents select from multiple teachers in these more complex, high-dimensional settings.
    \item \textbf{Adaptive Teachers.} Extending to teachers that co-evolve with the agent or to novel goals not seen during training would test robustness and scalability, building on our uncertainty ablation results ($\sigma \approx 1.0$) showing tolerance to imperfect goal knowledge.
    \item \textbf{Adaptive Risk Calibration.} Developing mechanisms for task-specific risk calibration---favouring safety in medical robotics versus higher risk in gaming---would provide principled design guidance for deploying adaptive and safe multi-teacher systems.
\end{itemize}

\subsection{Theoretical Connections to Risk Management} The connection to portfolio theory extends beyond mere analogy. Just as investors construct efficient frontiers to optimise risk-return trade-offs, learning agents implicitly navigate a similar frontier when selecting among teachers with varying reward-variance profiles. This connection suggests that established principles from financial optimisation might directly inform the design of teacher selection mechanisms.

Our explicit accuracy modelling reveals that advice quality dominates quantity, contrasting with budget-constrained teaching approaches that focus primarily on consultation frequency. The asymmetry between accuracy and availability effects indicates that trust, once established, requires minimal reinforcement, but once broken, becomes nearly impossible to restore. This finding has implications beyond multi-teacher learning, suggesting fundamental principles about trust dynamics in multi-agent systems.

\section{Conclusion}
\label{sec:conclusion}

This paper revealed that reinforcement learning agents systematically prefer conservative, low-reward teachers over those offering substantially higher rewards, challenging fundamental assumptions about reward maximisation. Through extensive experiments in multi-teacher learning environments, we demonstrated that agents develop risk-averse strategies prioritising consistency over optimality. The discovered phase transitions at critical thresholds provide practical guidelines for system design, while the persistence of conservative bias under concept drift validates its robustness within our experimental context. These findings suggest potential implications for human-robot and human-AI collaboration, indicating that effective training systems might benefit from incorporating teachers with varying risk profiles to leverage agents' observed tendency towards conservative selection. Future research should explore whether this bias emerges in continuous control tasks, investigate its manifestation in manipulation and multi-agent scenarios, and develop theoretical frameworks for predicting when conservative strategies dominate in multi-teacher systems.

\bibliographystyle{named}
\balance
\bibliography{bibliography}

\end{document}